%% file: main.tex
\documentclass[10pt,twocolumn,letterpaper]{article}

\usepackage[final]{cvpr}  

\usepackage{graphicx}
\usepackage{amsmath}
\usepackage{amssymb}
\usepackage{booktabs}
\usepackage{color}
\usepackage{subfiles}
\usepackage{mmstyles}
\usepackage{standalone}

\usepackage[pagebackref,breaklinks,colorlinks]{hyperref}

\usepackage[capitalize]{cleveref}
\crefname{section}{Sec.}{Secs.}
\Crefname{section}{Section}{Sections}
\Crefname{table}{Table}{Tables}
\crefname{table}{Tab.}{Tabs.}

\def\cvprPaperID{5252} 
\def\confName{CVPR}
\def\confYear{2022}

\def\fclip{\mathrm{CLIP}}

\begin{document}

\title{CLIP-GEN: Language-Free Training of a Text-to-Image Generator with CLIP}

\author{{Zihao Wang~~~ 
Wei Liu~~~ 
Qian He~~~ 
Xinglong Wu~~~
Zili Yi\footnote{Corresponding author.} ~~~} \\
ByteDance Inc.\\
{\tt\small \{wangzihao.vision, liuwei.jikun, heqian, wuxinglong, yizili\}@bytedance.com} \\
}

\twocolumn[{%
\renewcommand\twocolumn[1][]{#1}%
\maketitle
\begin{center}
\vspace{-0.25cm}
 \centering
\captionsetup{type=figure}
    \includegraphics[width=1.0\linewidth]{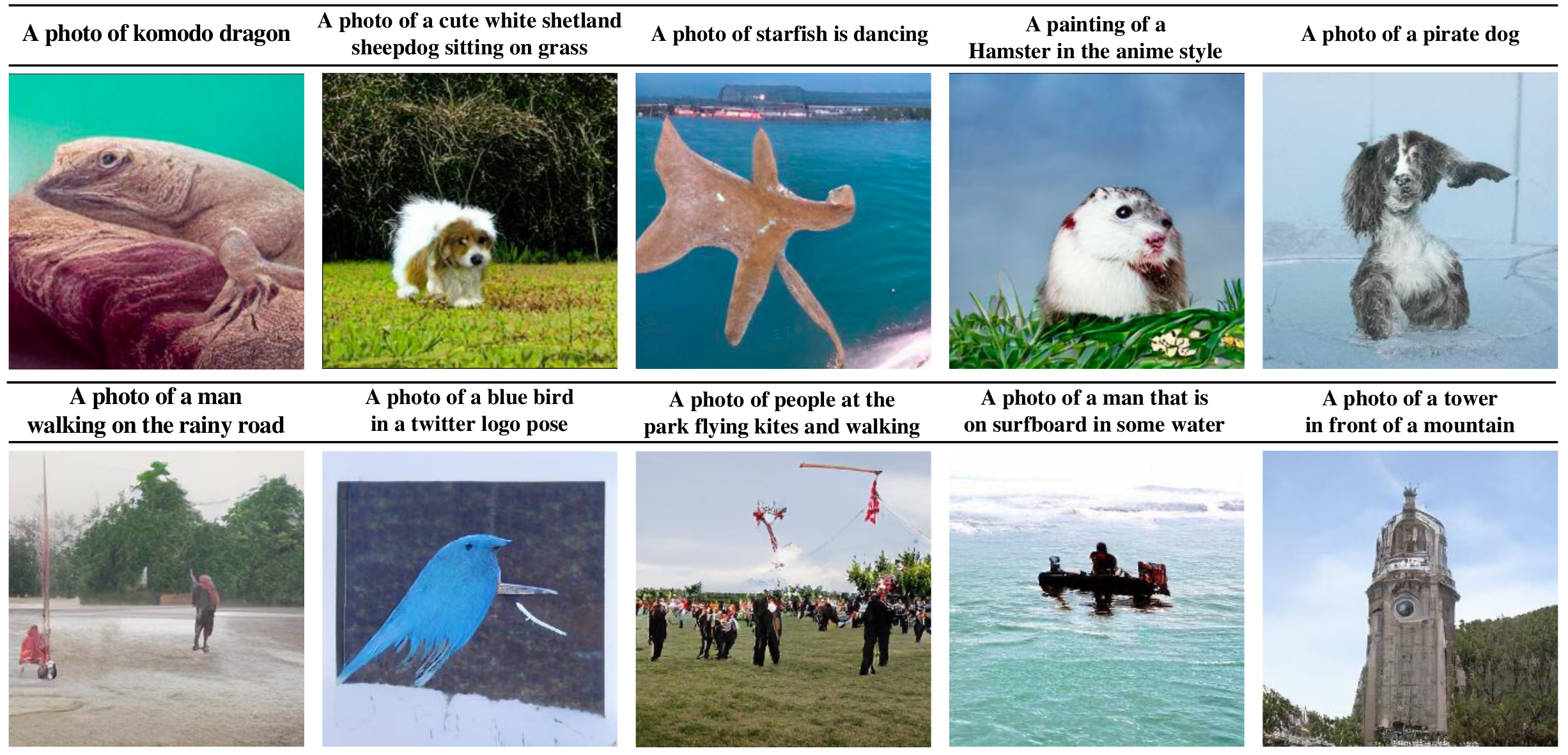}
    \caption{Exemplar images generated from text inputs by our method. The generator is only trained to reconstruct image tokens from image embeddings and has never seen any text inputs during training.\label{fig:teaser}}
\end{center}
}]

\begin{figure*}[bth]
   \centering
   \label{fig:atizzerts}
\end{figure*}

\begin{abstract}
Training a text-to-image generator in the general domain (e.g., Dall.e \cite{ramesh2021zero}, CogView \cite{ding2021cogview}) requires huge amounts of paired text-image data, which is too expensive to collect. In this paper, we propose a self-supervised scheme named as CLIP-GEN for general text-to-image generation with the language-image priors extracted with a pre-trained CLIP model \cite{radford2021learning}. In our approach, we only require a set of unlabeled images in the general domain to train a text-to-image generator. Specifically, given an image without text labels, we first extract the embedding of the image in the united language-vision embedding space with the image encoder of CLIP \cite{radford2021learning}. Next, we convert the image into a sequence of discrete tokens in the VQGAN codebook space \cite{esser2021taming} (the VQGAN model can be trained with the unlabeled image dataset in hand). Finally, we train an autoregressive transformer that maps the image tokens from its unified language-vision representation. Once trained, the transformer can generate coherent image tokens based on the text embedding extracted from the text encoder of CLIP upon an input text. Such a strategy enables us to train a strong and general text-to-image generator with large text-free image dataset such as ImageNet \cite{krizhevsky2012imagenet}. Qualitative and quantitative evaluations verify that our method significantly outperforms optimization-based text-to-image methods in terms of image quality while not compromising the text-image matching. Our method can even achieve comparable performance as flagship supervised models like CogView \cite{ding2021cogview}. \let\thefootnote\relax\footnotetext{*Corresponding author.}
\end{abstract}

\subfile{sections/introduction}

\subfile{sections/relatedwork}
\subfile{sections/method}

\subfile{sections/experiments}

\section{Conclusion}
In this paper, we propose the CLIP-GEN strategy to train a reliable and general text-to-image generator without using any paired text-image data, but with the image-language priors of CLIP and a set of unlabeled image data. Such strategy enables us to make use of the available huge text-free image dataset (e.g., ImageNet) to train a text-to-image generator as powerful as flagship models like CogView \cite{ding2021cogview} that is trained with huge amounts of paired data. The proposed strategy should see greater breakthroughs in the future if we use larger numbers of unlabeled images that are available on the Internet.

{\small
\bibliographystyle{ieee_fullname}
\bibliography{egbib}
}

\newpage

\begin{appendix}

\subfile{sections/supp}

\end{appendix}

\end{document}

%% file: sections/introduction.tex
\section{Introduction}
\label{sec:intro}
Text guided image generation in the general domain has been a challenging and frontier task in recent years. Early approaches (e.g., DMGAN \cite{zhu2019dm}, AttnGAN \cite{xu2018attngan}, DF-GAN \cite{tao2020df}, Obj-GAN  \cite{li2019object}, OPGAN \cite{li2019object}, SD-GAN \cite{yin2019semantics}, CPGAN \cite{liang2019cpgan}, XMC-GAN \cite{zhang2021cross}) that directly generate pixels from the given text embeddings with a convolutional generator have shown promising revolutions to generate images in limited domains. However, when designated to generate images in the general domain, these methods see poor results in terms of image quality and text-image matching.

Recently, transformer-based text-to-image generators such as DALL-E\cite{ramesh2021zero} and CogView~\cite{ding2021cogview} have achieved great progress. Such progress is owed to two factors. First, the discretized representations of images achieved by vector quantized models such as VQ-VAE~\cite{oord2017neural} and VQ-GAN~\cite{esser2021taming} enable an image to be represented in the same way as natural language, thus enabling a transformer to be trained upon the cross-modality text-image data in a unified framework. Second, the progress in terms of large model (consisting of tens or hundreds of billions of parameters) training significantly leverages the model capacities of modeling cross-modality data in general domains. So far, these large-transformer-based methods \cite{ramesh2021zero,ding2021cogview} achieve the best performance in terms of image quality, text-image relevance and range of domains. However, a limitation is that they require hundreds of millions of high-quality paired text-image data for the training, which are not publicly available and typically too expensive to acquire.

On the other hand, inspired by the recent progress in cross-modality language-vision pre-training and the unveiling of CLIP model \cite{radford2021learning}, various optimization-based methods attempt to search in the image space based on a query text by optimizing the text-image matching score of a pre-trained CLIP model. The image search domain used by these methods could be the latent codes of a pre-trained GAN model (e.g. BigGAN \cite{bigganclip2021}, StyleGAN \cite{patashnik2021styleclip,gal2021stylegan}, SWAGAN \cite{MAGnet2021}), the codebook of a VQGAN model \cite{vqganclip2021}, Diffusion Denoising Models \cite{diffusionclip2021}, structured representations such as a set of strokes (ClipDraw \cite{frans2021clipdraw}) or triangles \cite{triangleclip2021}, or a SIREN network \cite{DeepDaze2021,sirenfft2021} that maps spatial coordinates to pixels. These methods relieve the demands of huge paired datasets and computing resources. However, the images generated by methods of this stream are either limited to a specific domain (e.g., faces) or suffer low-quality (e.g., unnatural, structure-distorted or physically meaningless).


\begin{figure}[tb]
  \centering
   \includegraphics[width=1.0\linewidth]{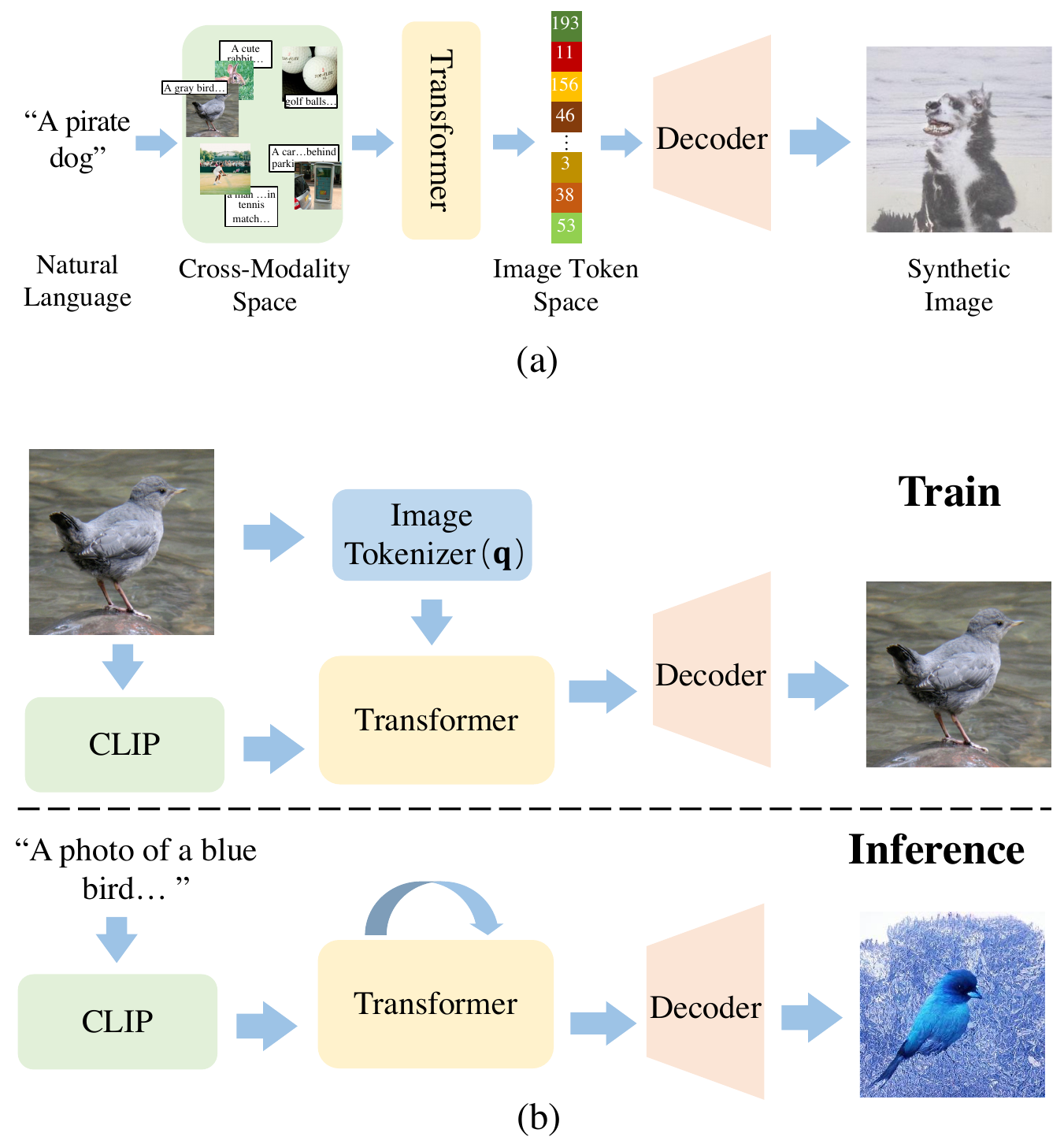}

   \caption{An overview of our purposed approach. (a) describes how our method maps a sentence to the corresponding image through the embedding space and the token space. (b) shows our training and testing pipeline. During training, the pre-trained CLIP model embeds the image to a cross-modality embedding and the pre-trained image tokenizer encodes the image into discrete image tokens. The autoregressive transformer learns to predict the image tokens with respect to the cross-modality embedding. During the inference, the CLIP model could either take an image or a sentence as the input, and then the transformer could predict coherent image tokens semantically related to the input.}
   \label{fig:intro}
\end{figure}

An analysis of methods of the two streams motivates us to seek a balance between the two. With the language-vision priors learned by the CLIP model, we shall be able to train a text-to-image generator without the use of any paired data. Considering the joint language-vision embedding space of a pre-trained CLIP is shared by both modalities, the image embedding extracted with the image encoder of the CLIP upon an image is also a good representation of textual semantics. If we train a transformer that maps an image embedding to the image itself, then the inference pipeline with text inputs is automatically bridged: starts from a text, goes though the united embedding space, and finally generates an image.

Specifically, we first extract the cross-modality embedding of the image in the joint language-vision embedding space with a pre-trained CLIP model \cite{radford2021learning}. Next, we convert the image into a sequence of discrete tokens in the VQGAN codebook space \cite{esser2021taming} which can be trained with the unlabeled image dataset in hand. Finally, an autoregressive transformer that predicts the image tokens based on its joint language-vision embedding is trained. During inference, the transformer can generate coherent image tokens based on the text embedding extracted upon an input text with the text encoder of CLIP, and the generated image tokens can be further reconstructed into an image with the VQGAN decoder: see Figure \ref{fig:intro}. Such a scheme relies on the assumption that the distribution of images embeddings used for training is well aligned with that of text embeddings during test, which shall hold if we intentionally ensure the statistical coherency of the semantic distribution of the training image data and text data used for test.

Contributions of this paper include:
\begin{itemize}
\item We propose a scheme to train a reliable and general text-to-image generator without any paired text-image data, but with a set of unlabeled images and a pre-trained CLIP model as prior. Our approach provides a promising new direction for high-fidelity text-to-image generation with accessible resources.

\item Qualitative and quantitative evaluations verify that our method outperforms optimization-based text-to-image methods (e.g., VQGAN+CLIP \cite{vqganclip2021}, BigGAN-CLIP \cite{bigganclip2021}) and CNN-based methods (DF-GAN \cite{tao2020df}) in terms of image quality while not compromising the text-image matching. Our model can even achieve comparable performance as the flagship supervised model like CogView \cite{ding2021cogview} which is trained with huge amounts of paired data.
\end{itemize}

%% file: sections/relatedwork.tex
\section{Related Work}
\noindent\textbf{Discrete Image Representation}

Oord et al. \cite{oord2017neural} first present an approach called Vector Quantized Variational Autoencoder (VQVAE) to learn discrete representations of images and model their distribution autoregressively with a convolutional architecture. \cite{razavi2019generating} extends this approach to use a hierarchy of learned representations to represent images of higher-resolution. \cite{esser2021taming} introduces self-attention layers \cite{ashish2017attn} to the bottleneck of the convolutional architecture with the expect to capture long-range interactions in high-resolution images, and adversarial training losses \cite{creswell2018generative} to enforce the learning of perceptually rich codebooks. In our approach, we use the perceptually rich discrete representations of images that preserve more photorealistic details and natural textures.

\noindent\textbf{Vision-Language Modelling}
Pre-training methods that have recently moved from raw text to multi-modal data (e.g., image-text) have revolutionized numerous multi-modal tasks (e.g., image-text matching \cite{tan2019lxmert,DBLPabs210300020,chen2020uniter}, image captioning \cite{zhang2021vinvl}, Visual Question Answering \cite{su2019vl,tan2019lxmert,li2020oscar,chen2020uniter}).  Cross-modality tasks require the understanding of both modalities, and the alignment and relationships between the two modalities. The pre-training enables the encoder to produce representations with fused cross-modality information, thus benefiting downstream tasks. 

The unveiling of CLIP model \cite{radford2021learning} is a big step for multi-modal pre-training. In the CLIP architecture, the image modality and the language modality are mapped with the image encoder and the text encoder respectively to the shared multi-modal embedding space. The text-image similarity score computed in the shared multi-modal embedding space of CLIP can serve as a metric of text-image alignment or an objective for text-guided image generation \cite{bigganclip2021,patashnik2021styleclip,gal2021stylegan,MAGnet2021,vqganclip2021,diffusionclip2021,frans2021clipdraw,triangleclip2021,DeepDaze2021,sirenfft2021}. However, in our strategy, we use the shared multi-modal embedding space in a creative way, i.e., training a reverse model that maps the shared embedding to the image modality.

\noindent\textbf{Text-to-Image Generation}
While there have been several attempts to improve the controllability of image generation by conditioning image synthesis on explainable priors (categories, attributes, label maps, edge maps, key points, depth-map), they often require users to follow some fixed control patterns. However, text-to-image generation that enables free-style user controls is a good choice, as natural language is easy to express and rich in information. 

Recent years see great progresses in this field and many methods have been proposed. The key differences of existing text-to-image approaches rely on what are used to represent texts (e.g., word embeddings) and images (e.g., GAN, VAE, VQVAE or raw pixels) respectively, and what model is used to bridge the two modalities. Early methods for text-to-image \cite{zhu2019dm,xu2018attngan,tao2020df,li2019object,li2019object,yin2019semantics,liang2019cpgan,zhang2021cross,zhang2017stackgan,qiao2019mirrorgan,zhang2018stackgan++} attempt to train a convolutional generator that predicts pixels directly from the given text embeddings. Recently, transformer-based generators \cite{ramesh2021zero,ding2021cogview} that map the textual embeddings to the discretized representations of images (VQGAN \cite{vqganclip2021} or VQVAE \cite{oord2017neural,razavi2019generating}) have achieved significantly better results than traditional CNN-based methods. Other streams rely on a pre-trained GAN model (e.g., StyleGAN \cite{karras2020analyzing,viazovetskyi2020stylegan2}) and attempt manipulate the style space based on textual inputs \cite{xia2021tedigan,bigganclip2021,patashnik2021styleclip,gal2021stylegan,MAGnet2021,rombach2020network}, or rely on a pre-trained text-image matching model (e.g., CLIP \cite{radford2021learning}) and attempt to optimize the image representations to satisfy the textual guidance \cite{vqganclip2021,diffusionclip2021,triangleclip2021,DeepDaze2021,sirenfft2021,bigganclip2021,patashnik2021styleclip,gal2021stylegan,MAGnet2021}. Our method takes the advantages of both transformer-based and CLIP-based methods. We make use of the knowledge priors learnt with CLIP and train a powerful transformer without any paired data.

%% file: sections/method.tex
\begin{figure*}[t]
  \centering
   \includegraphics[width=1.0\linewidth]{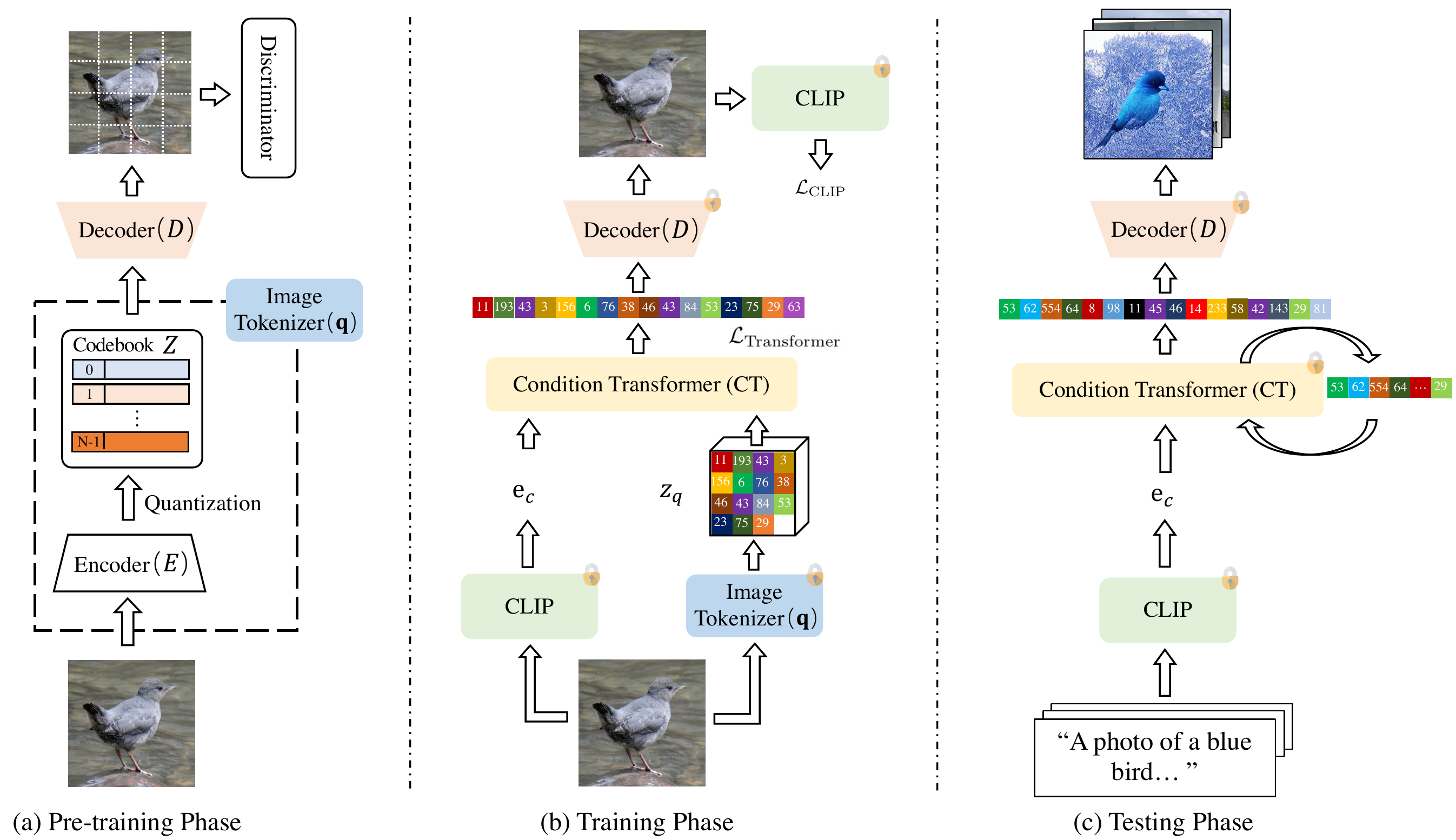}

   \caption{An illustration of our framework, the modules in the same color share the weights. In the pre-training phase, we train a vq-gan model as the image tokenizer. The weights of CLIP and image tokenizer are freezed after pre-training, we only optimize weights of the conditional transformer.}
   \label{fig:framework}
\end{figure*}

\section{Approach}
\label{sec:approach}

In this section, we introduce the details about our text-to-image generation framework and the training strategy.

As shown in \cref{fig:framework}, our model is made up of three components: a pre-trained language-image matching model (CLIP), an image tokenizer (VQ-GAN) and a conditional autoregressive transformer that takes
the image embedding $\ve_c$ of an image extracted from CLIP as the condition $\vc$, then generates the discrete image tokens $\vz$ of the same image.

\subsection{Language-Image Feature Extractor}

Contrastive Language–Image Pre-training (CLIP)~\cite{radford2021learning} has achieved great success in mapping the language-image inputs to a common embedding space. Given an image $I$ or a sentence $T$ as the input (denoted as $x$), the CLIP model can embed them into a common representation space: \begin{align}
\ve_c &= f_{\mathrm{CLIP}}(x), \mathrm{where}~x\in\{I, T\}. 
\end{align}

The official pre-trained CLIP model provided by~\cite{radford2021learning} is trained on 400-millions of text-image pairs with InfoNCE Loss:
\begin{align}
\cL_{\mathrm{InfoNCE}} &= \sum_{(I_{pos},T_{pos})}\log{\frac{\exp(\cos(I_{pos}, T_{pos}))}{\sum_{(I, T)}{\exp(\cos(I, T))}}}.
\end{align}
which learns robust representations of hetero-modality data to ensure the semantically relevant data to be close to each other in the common embedding space. 

Whereas it is too expensive to obtain large-scale and high-quality text-image pairs within our domain-of-interest, the CLIP model pre-trained on 400-million noisy pairs collected from Internet has shown enough capability to model language-vision data in general domains.
Specifically, we use the ViT-B/32 variation of CLIP~\cite{radford2021learning} in our experiments.

\subsection{Learning an Efficient Image Tokenizer}
\label{subsec:vqgan}

The recent VQ-VAE~\cite{oord2017neural} and VQ-GAN~\cite{esser2021taming} models have shown promising results to compress image patches into discrete image tokens. Such mechanisms enable images to be represented in the same way as natural language and easier to process with transformers. 

As shown in the pre-training phase of ~\cref{fig:framework}, we employ VQ-GAN to learn a perceptually rich codebook $\cZ=\{z\}^{K}\subset \Rbb^{dim_z}$ by optimizing all parameters of the encoder $E$, decoder $G$ and discriminator $D$. After the training finishes, the discriminator is removed. We only use the encoder $E$ and the codebook $\cZ$ as the image tokenizer $\vq(\cdot)$, and the decoder $G$ to reconstruct an image from its tokens.

Given an input image $I$, we first map it into a spatial embedding map of size $h\times w$,  $\hat{z} = E(I) \in \Rbb^{dim_z\times h\times w}$. Each embedding $\hat{z}_{ij}$ will then be hard-coded by looking up its nearest neighbor in the codebook:
\begin{align}
    z_q = \argmin_{z_k\in \cZ}{\lVert\hat{z}_{ij}-z_{k}\rVert}.
\end{align}
where the indices sequence $\{k\}^{h\times w}$ of $\{z_q\}$ is denoted as $\mathbf{s}$. 

The VQ-decoder is used to reconstruct an image from the token sequence $\mathbf{s}$, i.e., $\hat{I} = G(\mathbf{s})=G(\mathbf{\mathbf{q}(\mathbf{s}}))$. 

The VQGAN model can be optimized with an objective consisting of the reconstruction loss:
\begin{align}
\label{eq:L_vq}
    \cL_{\mathrm{vq}} &= \lVert I-\hat{I}\rVert ^2 + \lVert \mathrm{sg}[E(I)]-z_q \rVert_{2}^{2} + \lVert \mathrm{sg}[z_q]-E(I) \rVert_{2}^{2}
\end{align}
and the adversarial loss:
\begin{align}
\label{eq:L_D}
    \cL_{\mathrm{GAN}} &= [\log{D(I)} + \log(1-D(\hat{I})]
\end{align}
where $\mathrm{sg}(\cdot)$ is the stop-gradient operation.

\subsection{Conditional Autoregressive Transformer}
\label{subsec:autoreg}

The conditional autoregressive transformer is designated to predict image tokens based on its CLIP embedding. Given an input image $I$, we obtain its embedding with the CLIP image encoder $\ve_c = f_{\fclip(I)}$ and a row-major ordered sequence of image tokens $\textbf{s} = \vq(I) = \{s_1, s_2, ..., s_{h\times w}\}, s_i\in\{0, ..., |\cZ|\}$. Since the CLIP model only extracts high-level semantic information of an image, we expect the low-level image information of the image could be restored with the transformer in an autoregressive way, just as
\begin{align}
    p(\textbf{s}|\textbf{c}) &= \prod_{i}{p(s_i|s_{<i}, \ve_c)}
\end{align}

Once the complete set of tokens $\mathbf{s}$ are restored with respect to the image embedding, the pre-trained decoder $G$ could reconstruct the tokens back to an image, $\hat{I} = G(\textbf{s})$.


\subsection{Training Strategy}

\noindent We employ the two-stage training strategy.

\noindent \textbf{First Stage} We first train a VQ-GAN model with the image dataset in a self-supervised manner. As mentioned in \cref{subsec:vqgan}, all parameters of the encoder $E$, decoder $G$, codebook $\cZ$ and discriminator $D$ will be optimized during training. The training objective is:
\begin{align}
    \cL_{\mathrm{tokenize}} &= \argmin_{E, G, \cZ}\max_{D}{\Ebb_{x\sim p(x)}{[L_{VQ} + L_{GAN}]}}
\end{align}
where $\cL_{\mathrm{VQ}}$ and $\cL_{\mathrm{GAN}}$ are specified at ~\cref{eq:L_vq} and \cref{eq:L_D} respectively.

\noindent \textbf{Second Stage} The conditional autoregressive transformer is trained at this stage. Since we have paired input-output data (embedding$\rightarrow$image), our objective is a sum of the embedding reconstruction loss and a loss to maximize the likelihood of the corresponding image token.

The maximum-likelihood of the token sequence is enforce with 
\begin{align}
\cL_{\mathrm{Transformer}} = \Ebb_{x\sim p(x)}[-\log{p(s)}]
\end{align} 

To ensure the generated image can be mapped back to its embedding with the  CLIP image encoder, we employ the embedding reconstruction loss:
\begin{align}
\cL_{\mathrm{CLIP}} &= -\log{s(f_{\fclip}(G(\mathbf{s})), f_c)}
\end{align}

The training objective is the weighted combination of the two losses above:
\begin{align}
\cL &= \cL_{\mathrm{Transformer}} + \lambda \cL_{\mathrm{CLIP}}
\end{align}
where we set $\lambda=0.2$ in our implementation.

%% file: sections/experiments.tex
\begin{table*}[thb]
\begin{center}
\resizebox{\linewidth}{!}{%
\begin{tabular}{lcccccc}
\hline
\textbf{Experiment}  & Params  & GPT2 layers & GPT2 innder dim& Codebook embdim & Codebook size & Length of image tokens\\
\hline
ImageNet & 1.6B & 48 & 1536 & 256 & 16384 & 256 \\
COCO & 307M & 24 & 1024& 256 & 16384 & 256 \\
\hline
\end{tabular}
}
\end{center}
\caption{Hyper-parameters of our architecture in experiments.}
\label{tab:params}
\end{table*}

\section{Experiment}
\label{sec:exp}

In this section, we describe how we evaluate our method and compare with previous approaches. We first introduce the datasets used for training and validation and the implementation details of our approach on these datasets.
Then we make comprehensive comparisons between our method and previous text-to-image methods both quantitatively and qualitatively.

\begin{table}[htb]
\begin{center}
\resizebox{\linewidth}{!}{%
\begin{tabular}{lccccc}
\hline
\textbf{Model}  & \textbf{IS} $\uparrow$ & \textbf{FID-0} $\downarrow$  & \textbf{FID-1} $\downarrow$  &\textbf{FID-2} $\downarrow$ &\textbf{CapS} \\
\hline
AttnGAN~\cite{xu2018attngan} & 23.3 & 35.2 & 44.0 & 72.0 &  0.02763\\
DM-GAN~\cite{zhu2019dm} & \textbf{32.2} & 26.0 & 39.0 & 73.0 & 0.02801 \\
DF-GAN~\cite{tao2020df} & 18.7 & 26.0 & 33.8 & 55.9 &0.02802 \\
CogView~\cite{ding2021cogview} & 18.2 & 27.1 & 19.4 & \textbf{13.9} &\textbf{0.17403} \\
DALL-E~\cite{ramesh2021zero} & 17.9 & 27.5 & 28.0 & 45.5 & - \\
\hline
Ours & 21.4 & \textbf{20.7} & \textbf{18.5} & 17.4 &  0.13751 \\
\hline
\end{tabular}
}
\end{center}
\caption{Quantitative results on MS-COCO validation dataset.}
\label{tab:COCO}
\end{table}

\begin{figure*}[tbh]
  \centering
   \includegraphics[width=1.0\linewidth]{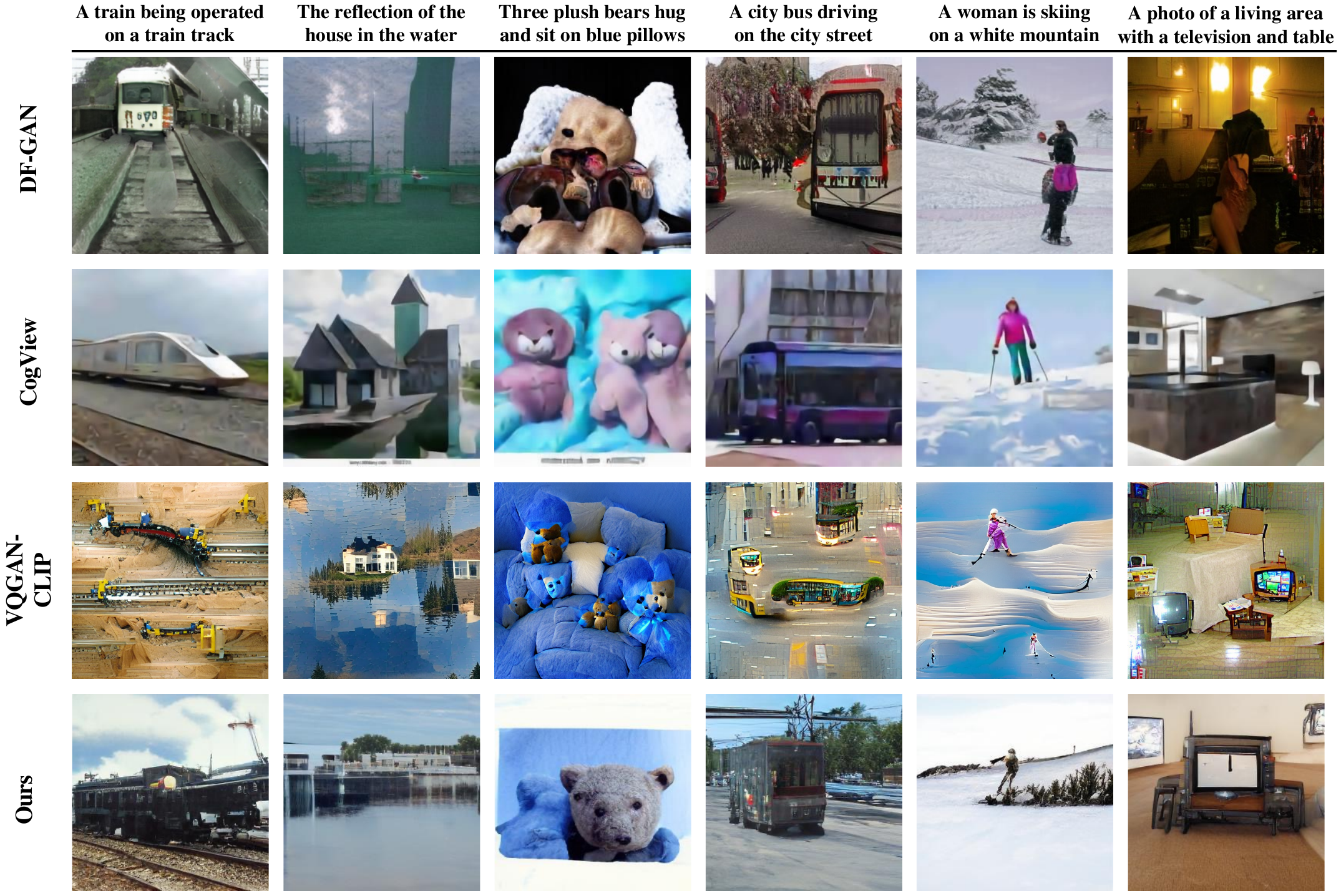}
   \vspace{0.4cm}
   \caption{Generated samples based on six textual descriptions from the MS-COCO validation dataset. Our results as demonstrated in this figure are generated with the model trained on only images from the complete MS-COCO dataset.}
   \label{fig:coco}
\end{figure*}


\begin{figure*}[tbh]
  \centering
   \includegraphics[width=1.0\linewidth]{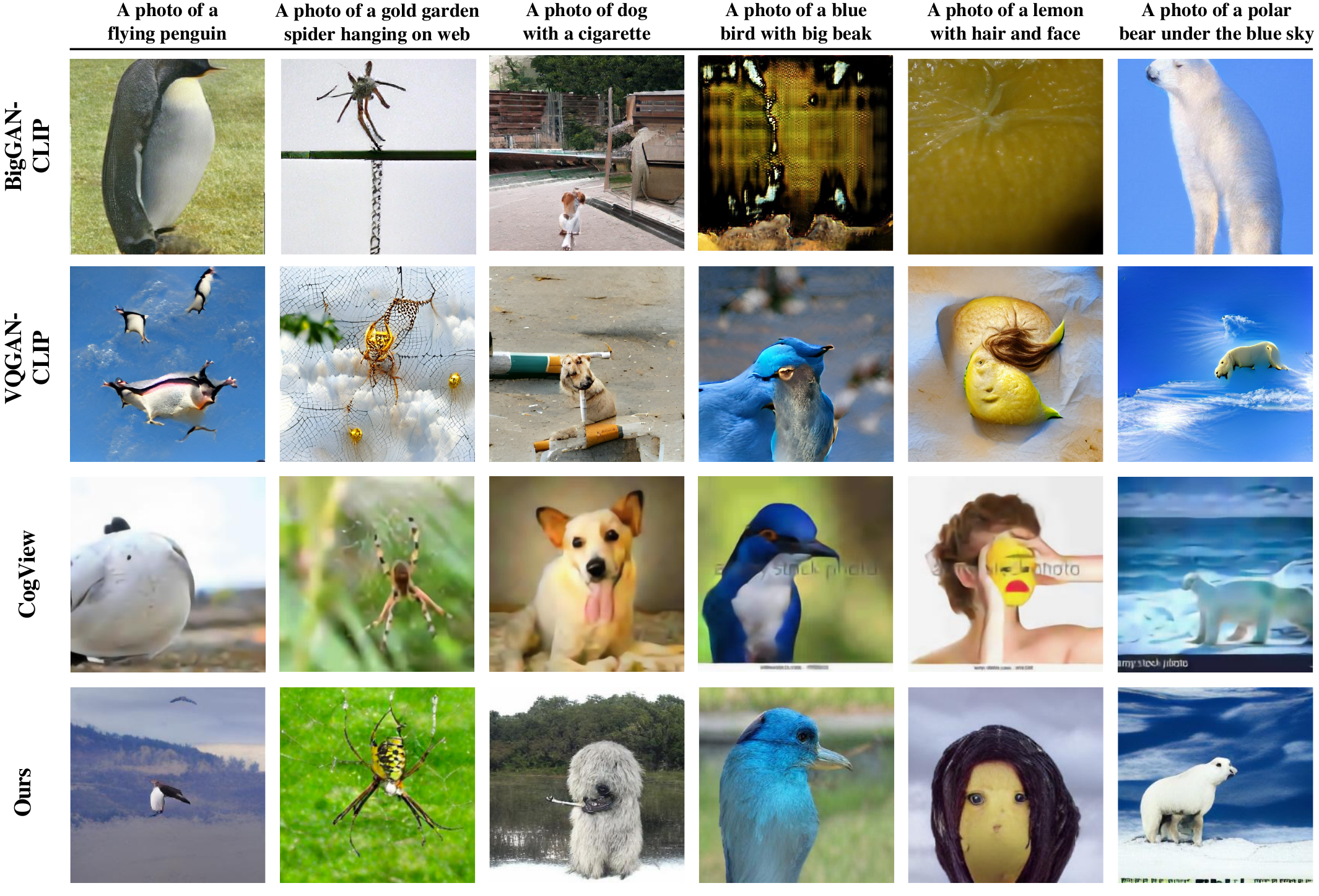}
   \vspace{0.4cm}
   \caption{Generated samples with different text-to-image methods. The captions are carefully composed so as to comply with the distribution of the ImageNet dataset, i.e., a visual description of an object covered by the ImageNet categorical labels.}
   \label{fig:imagenet}
\end{figure*}

\subsection{Datasets}
\label{subsec:datasets}
We train and evaluate our methods on two datasets: MS-COCO~\cite{lin2014microsoft} and ImageNet~\cite{krizhevsky2012imagenet}.

\noindent\textbf{MS-COCO}~is a widely used dataset for language-vision benchmarks. It contains 80k images for training and 40k test set images. Each image has 5 short sentence descriptions, which is not used in our method but used by competing methods (e.g., DM-GAN \cite{zhu2019dm}, DF-GAN \cite{tao2020df}, and AttnGAN~\cite{xu2018attngan}). We use the 2014 split of MS-COCO dataset in our experimental setting. We use the complete set of images to train the vQGAN and the text-to-image generator, while we only use the textual descriptions form the validation split for the quantitative evaluation and visual demonstration.  

\noindent\textbf{ImageNet}~has long been used to evaluate conditional generation tasks. It contains more than 14 million images, and a little more than 21 thousand groups or classes. We use the complete set of images to train the VQGAN and our text-to-image generator. For evaluations, we construct the input textual descriptions either by fitting the template of ``a photo of a [class name]'' ({\it{class name}} is an ImageNet category) or manually composing a caption like ``a photo of some [descriptive] objects with some [features]'' ({\it{descriptive}} can be some constraints of the color, size or other properties of the object, and {\it{features}} could be some accessories of the object).  

\label{subsec:quant}
\subsection{Implementation Details.}
For both datasets, we train a VQGAN with $|\cZ|=16384$ and $dim_z=256$. We use GPT2~\cite{radford2019language} as the architecture of our conditional transformer. We trained a 24-layers GPT2-medium for MS-COCO and a 48-layers GPT2-XL for ImageNet. The details of params are shown in ~\cref{tab:params}. The CLIP we used is the pre-trained ViT-B/32 model released by OpenAI \cite{radford2021learning}.

\subsection{Comparisons}
We compare our results with four existing approaches that are representative methods of different research streams (e.g., CNN-based methods, optimization-based methods and transformer-based methods). These methods are:

\noindent\textbf{DF-GAN}~\cite{tao2020df}, \textbf{DM-GAN}~\cite{zhu2019dm} and \textbf{AttnGAN}~\cite{xu2018attngan} represent the traditional approaches which use CNN generator to directly generate images with a textual condition. These methods provide pre-trained models on the MS-COCO dataset, so we can directly use those models for comparisons. 

\noindent\textbf{CogView}~\cite{ding2021cogview} is a flagship transformer-based method and serves as a good representation of fully-supervised and large transformer-based models~\cite{ding2021cogview, ramesh2021zero}, which are trained on tens of millions of high-quality text-image pairs. It achieves the best text-image relevance and FID metrics (see more details in \cite{ding2021cogview}), but the results by CogView suffer lack of perceptual details as images are encoded and decoded with VQ-VAE~\cite{radford2021learning}. Its pre-trained model is a large model with 4-Billion parameters trained on 50 million text-image pairs in the general domain and should cover the distribution of both ImageNet and MS-COCO images very well.

\noindent\textbf{VQGAN-CLIP}~\cite{vqganclip2021} represents for zero-shot opimization-based approach. It utilizes CLIP scores to guidance the optimization direction of latent codes of a pre-trained VQGAN model without any extra training. The CLIP model used in our comparisons is the pre-trained ViT-B/32 model released by OpenAI \cite{radford2021learning}. The VQGAN models for the two datasets respectively are the same as used in our method.

\noindent\textbf{BigGAN-CLIP}~\cite{bigganclip2021} Since DF-GAN cannot be trained on ImageNet (without text labels), we use the BigGAN-CLIP~\cite{bigganclip2021} as a substitute when conducting visual and quantitative comparisons on ImageNet dataset. Here the BigGAN model is the one pre-trained on ImageNet dataset and provided by \cite{brock2018large}.


\begin{figure*}[htb]
  \centering
   \includegraphics[width=1.0\linewidth]{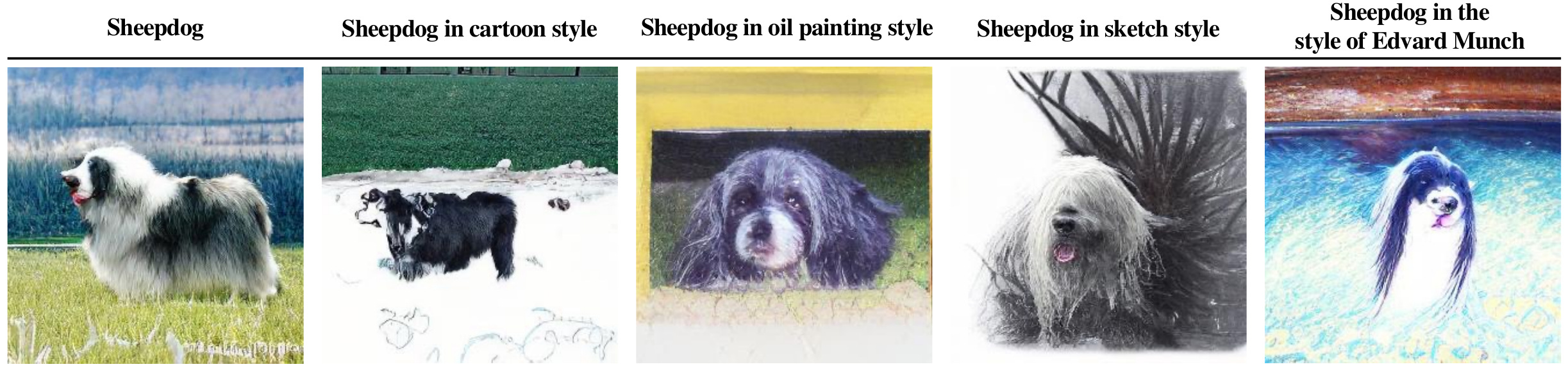}
   \vspace{0.25cm}
   \caption{The generated sheepdog pictures in different styles. These images are synthesized with our model trained on the ImageNet dataset without any data augmentation.}
   \label{fig:style}
\end{figure*}


\subsection{Quantitative Results}

\noindent\textbf{Evaluation Metrics}
To evaluate the quality of generated images, we use the stantard metrics as in \cite{ding2021cogview}: Inception Score (IS)~\cite{tim2016is}, Fr´echet Inception Distance (FID)~\cite{martin2017fid} and CapS \cite{ding2021cogview}. IS calculates KL-divergence between conditional distribution and marginal distribution given an image classifier. FID computes the Fr´echet distance between the distribution of Inception features of synthetic images and real-world images. CapS measures the semantic similarities between the input text and the generated image. The quantitative results are computed over 30,000 images generated based on diverse validation captions and 30,000 ground-truth images related to the captions.

\noindent\textbf{MS-COCO}~\cref{tab:COCO} shows the comparison between our method and previous methods on text-guided scene image synthesis. Our method achieves the best FID-0 and FID-1 due to the perceptually rich results generated by VQGAN and coherent image structures. The CapS score is lower than CogView by 4\% but significantly better than other competing methods.

\noindent\textbf{ImageNet.}~\cref{tab:IN} shows the class-conditional generation results on ImageNet. We compared our method with similar methods based on discrete image tokens. Since the models require text inputs, we use the textual prompting (\textit{A photo of a \{ImageNet label\}.}) as the input. Our method achieves the best FID metrics, implying that our method achieves better image quality and more coherent semantic distribution.
\begin{table}[htb]
\begin{center}
\resizebox{\linewidth}{!}{%
\begin{tabular}{lcc}
\hline
\textbf{Model}  & \textbf{IS} $\uparrow$ & \textbf{FID} $\downarrow$ \\
\hline
VQGAN+CLIP~\cite{vqganclip2021} & 20.8 & 77.0 \\
VQVAE-2~\cite{razavi2019generating} & $\sim$45 & $\sim$31  \\
VQGAN+label conditioned Transformer~\cite{esser2021taming} & \textbf{70.6} & 17.0 \\
\hline
Ours & 45.16 & \textbf{16.74}\\
\hline
\end{tabular}
}
\end{center}
\caption{Quantitative results on ImageNet dataset.}
\label{tab:IN}
\end{table}

\subsection{Qualitative Results}
\label{subsec:examples}

As shown in Fig. \ref{fig:coco} and Fig. \ref{fig:imagenet}, compared to the competing methods, our method could generate high-fidelity images with more details. Generally speaking, the results of VQGAN-CLIP \cite{vqganclip2021} are non-realistic and suffer severe image distortion. CogView \cite{ding2021cogview} can generate good image structures but fails to produce realistic textures as they use the VQVAE \cite{oord2017neural} to discretize images. The BigGAN-CLIP \cite{bigganclip2021} method sees more natural details than VQGAN-CLIP but suffer distorted image structures either.  The DF-GAN \cite{tao2020df} can generate acceptable image structures with perceptually rich details but is prone to producing local regional artifacts. The visual evaluation matches the quantitative results well.

As shown in Fig. \ref{fig:coco}, our model can successfully capture the semantic concepts such as ``reflection in the water'' ($2^{nd}$ column), but fails to match the numeric concepts like ``three plush bears'' ($3^{rd}$ column). These concepts are captured by CogView model \cite{ding2021cogview} very well.


To examine the generalization ability of our method, we attempt to generate images under out-of-distribution language descriptions. As shown in  ~\cref{fig:imagenet}, some of the descriptions (e.g. ``a dog with a cigarrete'', ``a lemon with hair and face'') do not even have a corresponding real-world image. Our model that is trained on the realistic images can surprisingly generate images well-aligned with these out-of-distribution texts. However, CogView \cite{ding2021cogview} that is trained upon amounts of image with textual labels fails to match those decorative words (e.g., ``flying'', ``with a cigarrete'', ``with big beak'').

We also explore the generalization ability of our method in terms of stylized synthesis. We attempt to generate images under special style descriptions (e.g., ``sketch'', ``oil painting'' or even ``style of Edvard Munch''). As shown in ~\cref{fig:style}, our model can successfully synthesize stylized pictures even without seeing many stylized training samples as no style augmentation is applied during training.

%% file: sections/supp.tex

\twocolumn[{%
\renewcommand\twocolumn[1][]{#1}%
\begin{center}
     {\Large \bf CLIP-GEN: Language-Free Training of a Text-to-Image Generator with CLIP \\ \rule{2cm}{0.7pt} \\Supplementary Material \par}
      {
      \large
      \lineskip .5em
      \par
      }
      \vskip .5em
      \vspace*{12pt}
  \end{center}
}]


In this supplementary material, we provide additional implementation details, as well as some additional visual results.

In~\cref{sec:implementation}, we present the architecture and hyperparameters we used during training. In ~\cref{sec:res}, we show additional generated images with different sample strategies, and some generated illustrations for famous lines of poetry.

\subfile{supp_sections/implementation.tex}

\subfile{supp_sections/results.tex}

%% file: sections/supp_sections/implementation.tex
\section{Implementation Details}
\label{sec:implementation}

\noindent\textbf{Pre-trained VQ-GAN} In the pre-training stage, we train two VQ-GAN~\cite{esser2021taming} models on ImageNet~\cite{krizhevsky2012imagenet} and COCO~\cite{lin2014microsoft} dataset respectively. Both models have the same architecture and hyperparameters. Specifically, we set the codebook size $|\cZ|=16384$, embedding size $dim_z=256$, compression factor $f=16$ and sequence length $|s|=16\times 16=256$.

Following the implementation of the original VQ-GAN, we use a 3-layer PatchGAN~\cite{isola2017image} discriminator with perception loss. The discriminator is not trained before 25,000 steps. The models are trained with Adam optimizer with the initial learning rate $0.0000045$. During training, input images are resize and random cropped to $256\times 256$ while the model reconstruct the image in the same size. The model is finally trained for 100 epochs with a batch size of 16.

\noindent\textbf{Autoregressive Transformer} In the second stage, we only trained the autoregressive transformer. The weights of VQ-GAN and CLIP~\cite{radford2021learning} are fixed. We choose the GPT-2~\cite{radford2019language} architecture in our implementation.

On COCO dataset, we use the medium model with 24 layers, 16 attention heads and the inner dim is 1024. On ImageNet dataset, we use the modified XL model with 48 layers, 24 attention head and the inner dim is 1536. The feature extracted from CLIP is liner probed to the same number of inner dim before input to the transformer.

We train the model with Adam optimizer and the initial learning rate is set to $0.00001$. The medium model is trained with a batch size of 32 for 1 million steps and the XL model is trained with a batch size of 16 for 2 million steps.

%% file: sections/supp_sections/results.tex
\section{Additional Results}
\label{sec:res}
\begin{figure*}[htb]
  \centering
   \includegraphics[width=1.0\linewidth]{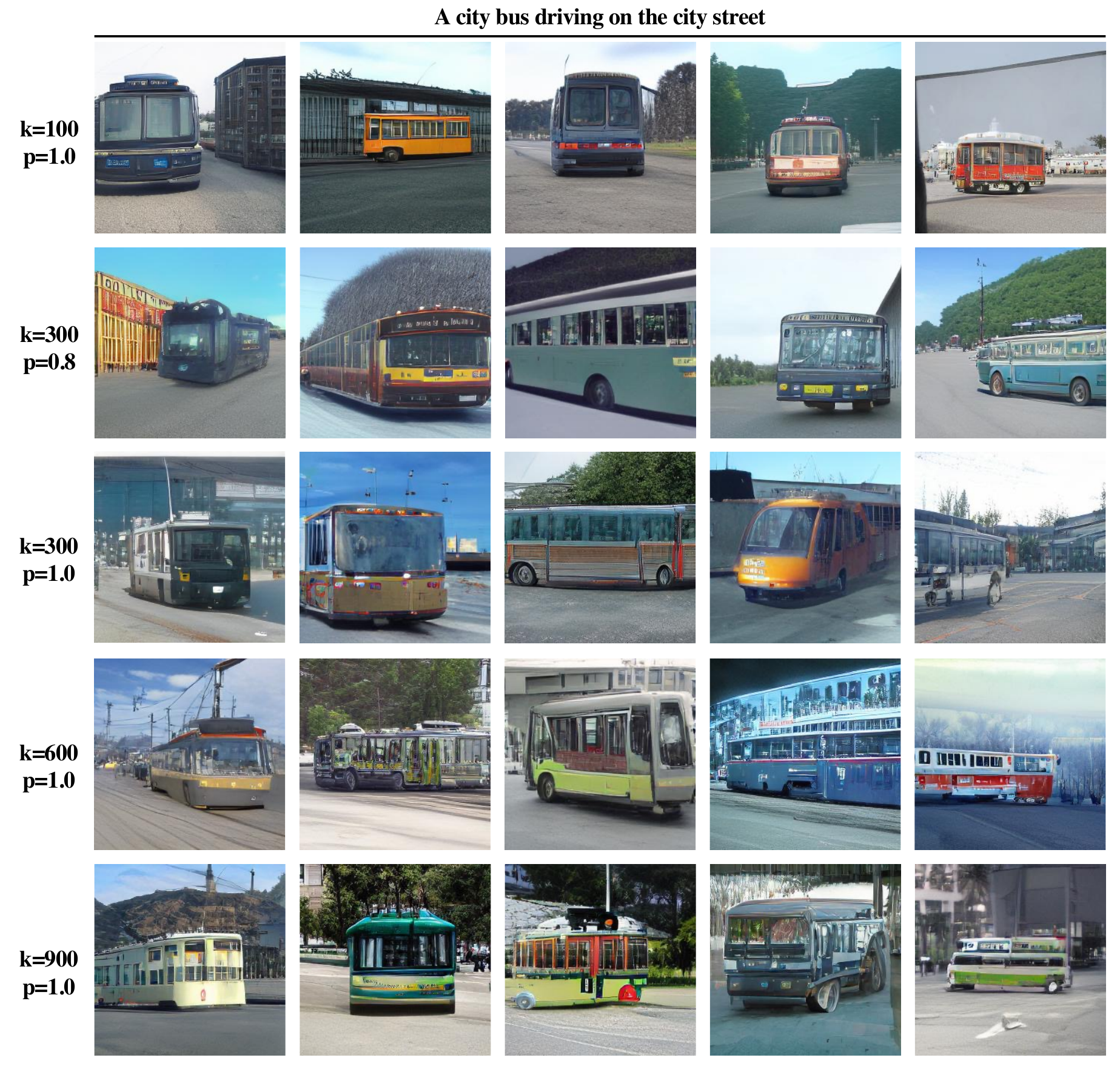}
   \caption{The generated results with different sample strategies.}
   \label{fig:sample1}
\end{figure*}

\begin{figure*}[htb]
  \centering
   \includegraphics[width=1.0\linewidth]{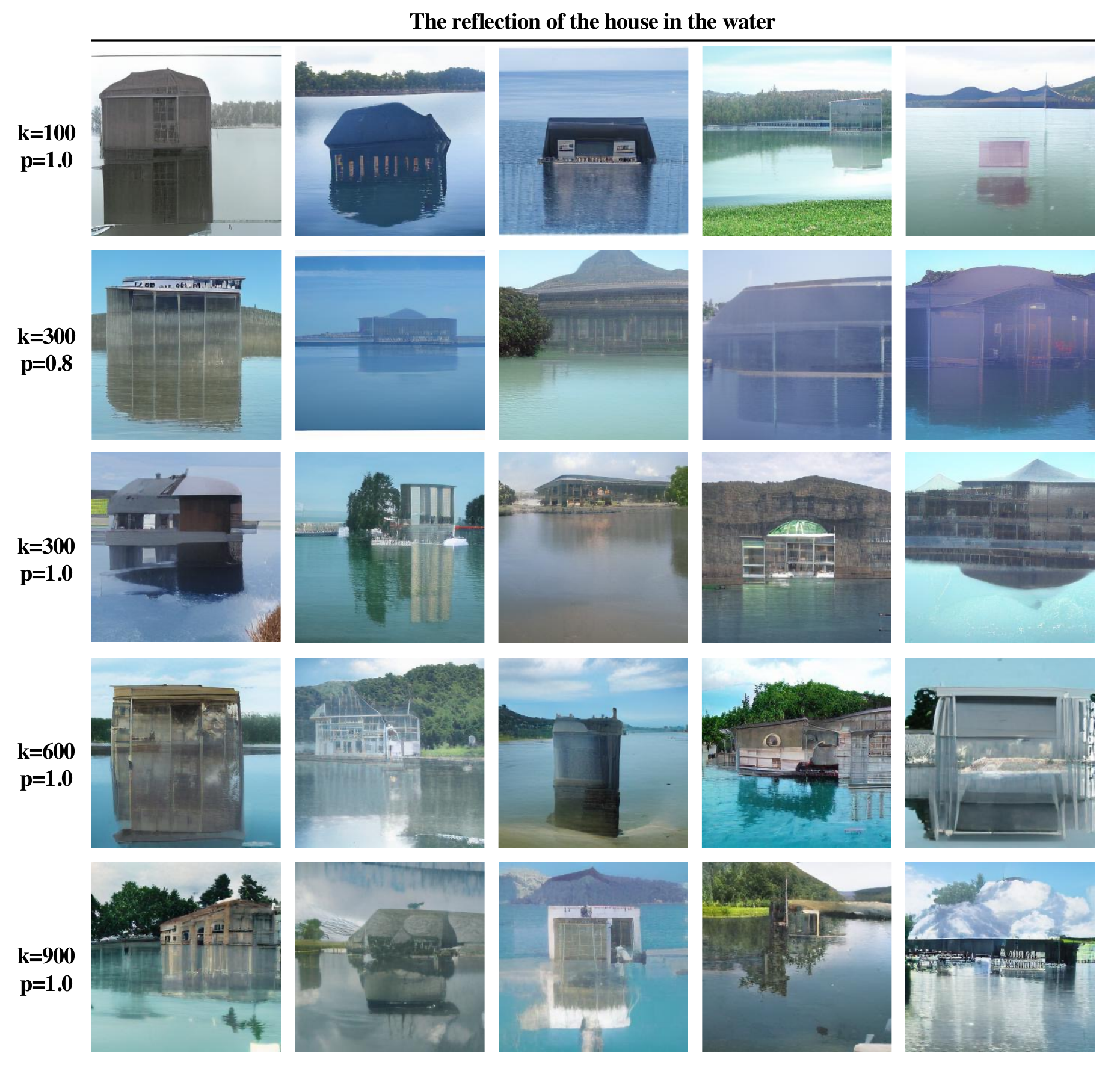}
   \caption{The generated results with different sample strategies.}
   \label{fig:sample2}
\end{figure*}

\begin{figure*}[htb]
  \centering
   \includegraphics[width=1.0\linewidth]{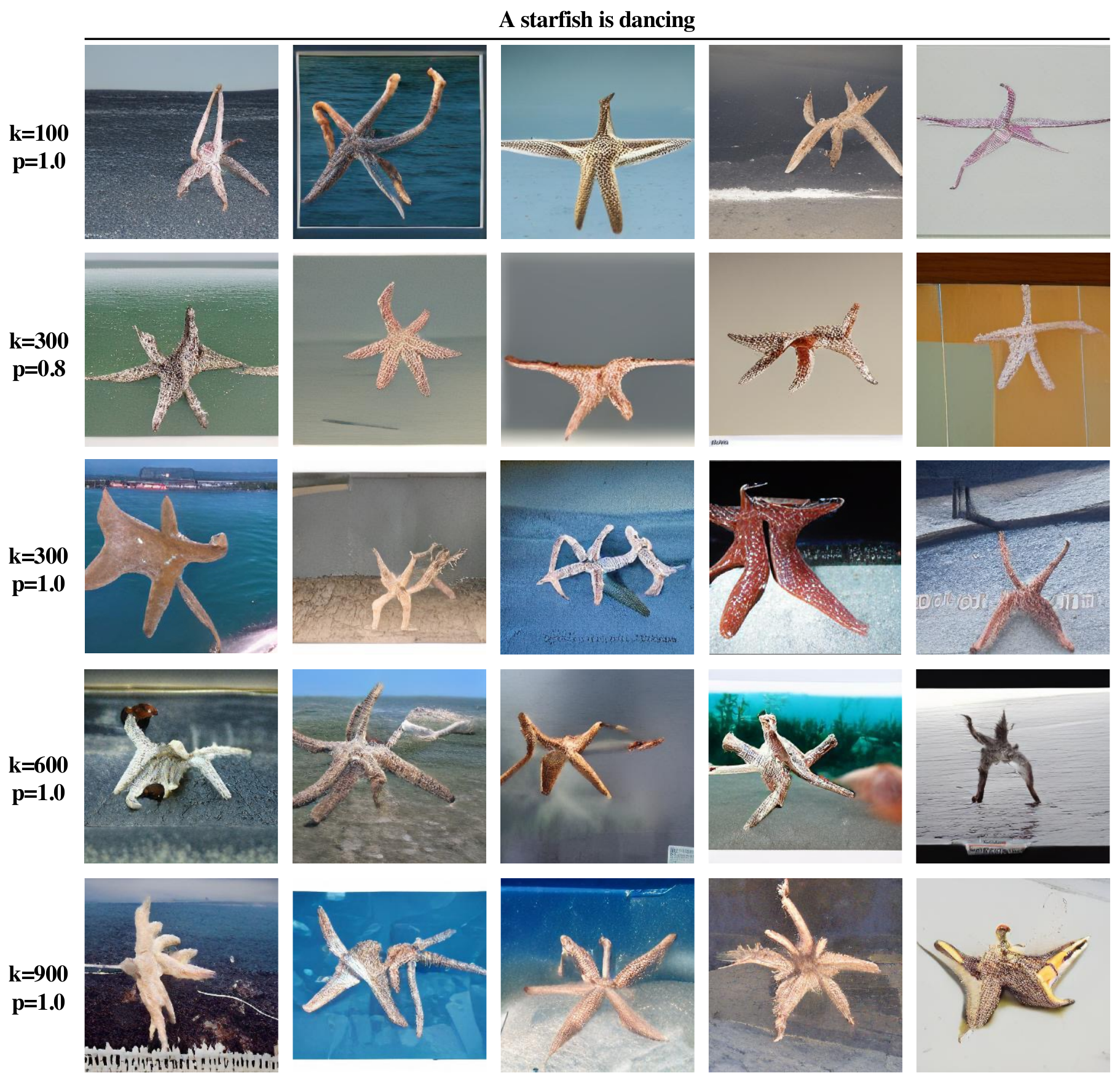}
   \caption{The generated results with different sample strategies.}
   \label{fig:sample3}
\end{figure*}

\noindent\textbf{Different Sample Strategies}
We employ top-k/p sample strategy to predict the image tokens with the autoregrresive model.
The top-k sampling only retains the highest $k$ logits.
The top-p sampling first normalize the logits into a probability distribution with softmax then remove the tokens with cumulative probability above the given threshold $p$. We will show our model's generated images with different tuned $k$ and $p$ in ~\cref{fig:sample1,fig:sample2,fig:sample3}. It shows that the small $k$ and $p$ will only retrain a few token candidates and make the generated images be simple and naive, while sampling with large $k$ and $p$ leads to more diverse but chaos results. We found $k=300, p=1.0$ could be a good trade off between diversity and quality, the images shown in our paper are sampled with that hyper-parameters.

\begin{figure*}[tb]
  \centering
   \includegraphics[width=1.0\linewidth]{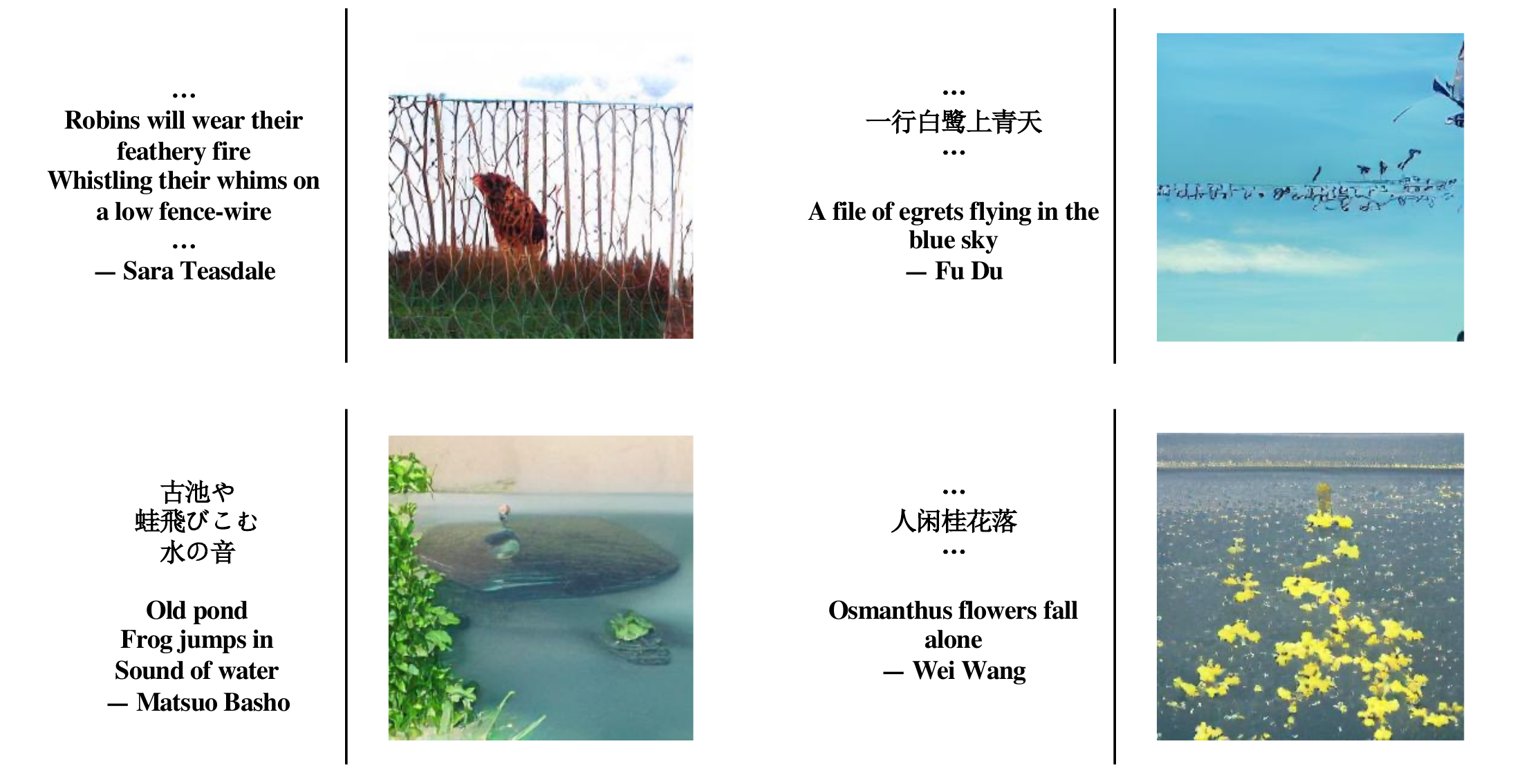}
   \caption{Generated illustrations for some famous lines of poetry.}
   \label{fig:poem}
\end{figure*}

\noindent\textbf{Illustration Generation}
To better show the model's capability, we selects several famous lines of poetry as the input and shows the generated painted style illustrations in~\cref{fig:poem} 

